\pdfoutput=1
\documentclass[11pt]{article}

\usepackage{acl}

\usepackage{times}
\usepackage{latexsym}
\usepackage{inconsolata}
\usepackage{amsmath}
\usepackage{amsfonts}
\usepackage{amssymb}
\usepackage{graphicx}
\usepackage{multirow}
\usepackage{booktabs}
\usepackage{diagbox}
\usepackage{bbding}
\usepackage{enumitem}
\usepackage{commath}
\usepackage{algorithm}
\usepackage{algorithmic}
\usepackage[toc,page]{appendix}
\usepackage{csquotes}
\usepackage{tabularx}
\usepackage{listings}
\usepackage[T1]{fontenc}
\usepackage[utf8]{inputenc}

\usepackage{microtype}

\title{Set-Aligning Framework for \\ Auto-Regressive Event Temporal Graph Generation}

\author{Xingwei Tan$^{1,2}$, Yuxiang Zhou$^2$, Gabriele Pergola$^1$, Yulan He$^{1,2,3}$ \\
  $^1$Department of Computer Science, University of Warwick, UK\\
  $^2$Department of Informatics, King's College London, UK\\
  $^3$The Alan Turing Institute, UK\\
  \texttt{\{Xingwei.Tan, Gabriele.Pergola.1\}@warwick.ac.uk}\\
  \texttt{\{Yuxiang.Zhou, Yulan.He\}@kcl.ac.uk}\\}

\begin{document}
\maketitle
\begin{abstract}
Event temporal graphs have been shown as convenient and effective representations of complex temporal relations between events in text.
Recent studies, which employ pre-trained language models to auto-regressively generate linearised graphs for constructing event temporal graphs, have shown promising results.
However, these methods have often led to suboptimal graph generation as the linearised graphs exhibit set characteristics which are instead treated sequentially by language models.
This discrepancy stems from the conventional text generation objectives, leading to erroneous penalisation of correct predictions caused by the misalignment of elements in target sequences.
To address these challenges, we reframe the task as a conditional set generation problem, proposing a Set-aligning Framework tailored for the effective utilisation of Large Language Models (LLMs).
The framework incorporates data augmentations and set-property regularisations designed to alleviate text generation loss penalties associated with the linearised graph edge sequences, thus encouraging the generation of more relation edges.
Experimental results show that our framework surpasses existing baselines for event temporal graph generation. 
Furthermore, under zero-shot settings, the structural knowledge introduced through our framework notably improves model generalisation, particularly when the training examples available are limited.\footnote{For access to experimental code and data, please refer to: \href{https://github.com/Xingwei-Warwick/Set-Aligning-Event-Temporal-Graph-Generation}{https://github.com/Xingwei-Warwick/Set-Aligning-Event-Temporal-Graph-Generation}}
\end{abstract}

%%%%%%%%%%%%%%%%%%%%%%%%%%%%%%%%%%%%%%%%%%%%%%%%%%%%%%%%%%%%%%%%%%%%%%%%%%%%%%%
%%%%%%%%%%%%%%%%%%%%%%       Introduction         %%%%%%%%%%%%%%%%%%%%%%%%%%%%%
%%%%%%%%%%%%%%%%%%%%%%%%%%%%%%%%%%%%%%%%%%%%%%%%%%%%%%%%%%%%%%%%%%%%%%%%%%%%%%%
\section{Introduction}

Understanding the temporal relation between events mentioned in long documents is crucial to modelling complex text with articulated narratives.
One of the widely adopted benchmarks for event temporal relation understanding is the SemEval 2013 TempEval-3~\cite{uzzaman-etal-2013-semeval}, requiring end-to-end generation of event temporal graphs directly from raw text.
An event temporal graph is a natural representation of temporal information, with the nodes representing events and the edges the temporal relationships between them, such as ``\emph{before}'', ``\emph{after}'', or ``\emph{simultaneous}''.

Existing studies typically approach the problem of constructing event temporal graphs through a two-step pipeline, with the first step focusing on detecting events in text, and the second step on classifying the temporal relations between them \cite{mcdowell-etal-2017-event,ning-etal-2018-cogcomptime}.
However, such pipeline-based approaches suffer from well-known limitations, including (i) the need for fine-grained annotations at each step; and (ii) the potential for error propagation throughout the pipeline. 
In the first step, the event extractor aims at locating as many event triggers as possible in the given documents, leading to the inclusion of numerous trivial events that often lack relevance to the narrative and have no relation with other events. 
As a result, the next step for temporal relational extraction becomes burdened with many noisy events, significantly impacting the overall accuracy and efficiency of the models.

To address these limitations, \citet{madaan-yang-2021-neural} introduced a reformulation of the task by generating event temporal graphs directly through conditional text generation. 
This approach allows for the use of pre-trained language models and, more importantly, overcomes the typical limitations associated with the pipeline architecture. 
While this method involved fine-tuning a text generation model, such as GPT-2, for the generation of linearised event temporal graphs as sequences, it fails to consider an important aspect. 
Specifically, it does not account for the fact that the target sequence (i.e. the list of event temporal relations) is order-invariant, and should therefore be treated as a \emph{set} rather than as an ordered sequence.
For example, the following two sequences represent the same temporal graph:

{\scriptsize
\begin{align*}
    &\texttt{S1: [(Cuomo leaving his office, before, speak to reporters),} \\
    &\hspace{0.5cm}\cdots\texttt{(Cuomo leaving, before, met with representatives)]}\\
    &\texttt{S2: [(Cuomo leaving, before, met with representatives),} \\
    &\hspace{0.5cm}\cdots\texttt{(Cuomo leaving his office, before, speak to reporters)]}
\end{align*}
}

\noindent In this scenario, the conventional text generation loss will (mistakenly) yield a high value because most of the tokens in the corresponding positions do not match, even though the event relations are the same.
This issue has a detrimental effect on the model performance for several reasons. First, it discourages the language model from generating additional edges. Generating more edges implies a greater number of potential permutations in the edge sets, making it less likely to match the target.
Secondly, if the initially generated edge in the sequence differs in token count from the one in the target, it causes all subsequent edges to misalign with the target, even if they are identical, leading to a high loss value.

In this work, we propose a Set-Aligning Framework (SAF) that enables efficient employment of LLMs for auto-regressive event temporal graph generation. 
SAF incorporates a group of novel regularisations, named Set Property Regularisations (SPR), along with augmented data, which aims at tackling the problems associated with the use of LM loss in contextualised graph generation by mitigating its penalisation towards the target sequences.
For example, the S1 and S2 above are different sequences of the same edge set.
Even if S1 has the same order as the target edge sequence and thus has a lower LM loss than S2, both of them will be added with the same SPR.
Therefore, the relative difference of their loss values becomes smaller, which avoids overfitting the model towards the specific edge order of S1.
Moreover, if the model explores generating one more edge after S2 and the edge is correct, the SPR value will decrease while the LM loss will probably increase.

Using the proposed SAF, we fine-tune language models from the T5 \cite{JMLR:v21:20-074} family with weak supervision.
Additionally, we introduce the first human-annotated dataset for contextualised event temporal graph generation built on the New York Times, which we combine with existing event relation extraction datasets to evaluate the effectiveness of the SAF framework.
Experiments on the newly annotated New York Times corpus\footnote{\url{https://doi.org/10.35111/77ba-9x74}} show that SAF significantly increases the number of generated edges, resulting in improved recall.
Furthermore, we assess the performance of our approach on existing sentence-level event temporal relation extraction datasets, namely MATRES \cite{ning-etal-2018-multi} and TB-Dense \cite{cassidy-etal-2014-annotation}, under zero-shot settings, and we find that the structural knowledge introduced through the proposed SAF has an even greater impact on model generalisation when the training examples available are limited.

Our contributions are three-folded:
\begin{itemize}
\item We introduce a model-agnostic framework, called SAF, for event temporal graph generation. SAF incorporates novel Set-Aligning regularisations, data augmentation, and weak supervision techniques.
\item We offer a human-annotated test set and a weakly-supervised dataset specifically designed for document-level event temporal generation.
\item Our extensive experimental results in various settings demonstrate the effectiveness of our proposed model.
Our thorough analysis shows that our SAF framework encourages language models to generate at least 24\% more edges than previous graph generation approaches across various datasets.
\end{itemize}

%%%%%%%%%%%%%%%%%%%%%%%%%%%%%%%%%%%%%%%%%%%%%%%%%%%%%%%%%%%%%%%%%%%%%%%%%%%%%%%
%%%%%%%%%%%%%%%%%%%%%%%%%   Related Work         %%%%%%%%%%%%%%%%%%%%%%%%%%%%%%
%%%%%%%%%%%%%%%%%%%%%%%%%%%%%%%%%%%%%%%%%%%%%%%%%%%%%%%%%%%%%%%%%%%%%%%%%%%%%%%
\section{Related Work}
% We discuss the connection between our study and three lines of prior research.

\subsection{Event Temporal Graph}
The task of event temporal graph extraction serves as an important task for evaluating an end-to-end system which takes raw text as input and output TimeML annotations (i.e., temporal relations) \cite{uzzaman-etal-2013-semeval}.
Early attempts on the task include CAEVO \cite{mcdowell-etal-2017-event} and Cogcomptime \cite{ning-etal-2018-cogcomptime}, which relied on a combination of statistical and rule-based methods.
In recent years, more efforts have been put into developing specialised sub-systems with neural network-based approaches \cite{ning-etal-2019-improved,han-etal-2019-deep,tan-etal-2021-extracting}.
The emergence of large language models has paved the way for end-to-end learning, treating temporal graph generation as conditional text generation \cite{madaan-yang-2021-neural}.
To tackle the set misalignment issue which remained unexplored in \citet{madaan-yang-2021-neural}, we propose a framework based on a group of novel Regularisations, aiming at enhancing autoregressive event temporal graph generation.

It is worth noting that there is another related and more widely-recognised task called \emph{temporal relation extraction}, which aims at classifying the type of temporal links between pre-extracted events \cite{wang-etal-2020-joint,wen-ji-2021-utilizing,tan-etal-2023-event}.
While \citet{han-etal-2019-joint} proposed a joint extraction model for events and event temporal relations, they rely on event extraction supervision signals, which our work does not need.

\subsection{Graph Generation with Language Models}
Generating graphs with language models has been explored in many areas.
For example, \citet{bosselut-etal-2019-comet} fine-tunes GPT on the ATOMIC commonsense knowledge graph \cite{10.1609/aaai.v33i01.33013027}.
\citet{melnyk-etal-2022-knowledge} proposed a multi-stage system for knowledge generation based on T5.
However, these studies do not generate an entire graph in one generation.
In contrast, \citet{madaan-etal-2021-think} generated inference graphs using a combination of a graph generator and a graph corrector for queries in defeasible reasoning.
\citet{zaratiana2023an} generate entities and entity relations with an auto-regressive LM, but they did not consider the set property of the target.
Different from them, we focus on the set property of the generation sequence, which is particularly important in the setting where both the input document and output sequence are considerably longer.

\begin{figure*}[t] % 'h' for here
   \centering
   \includegraphics[width=1.0\textwidth]{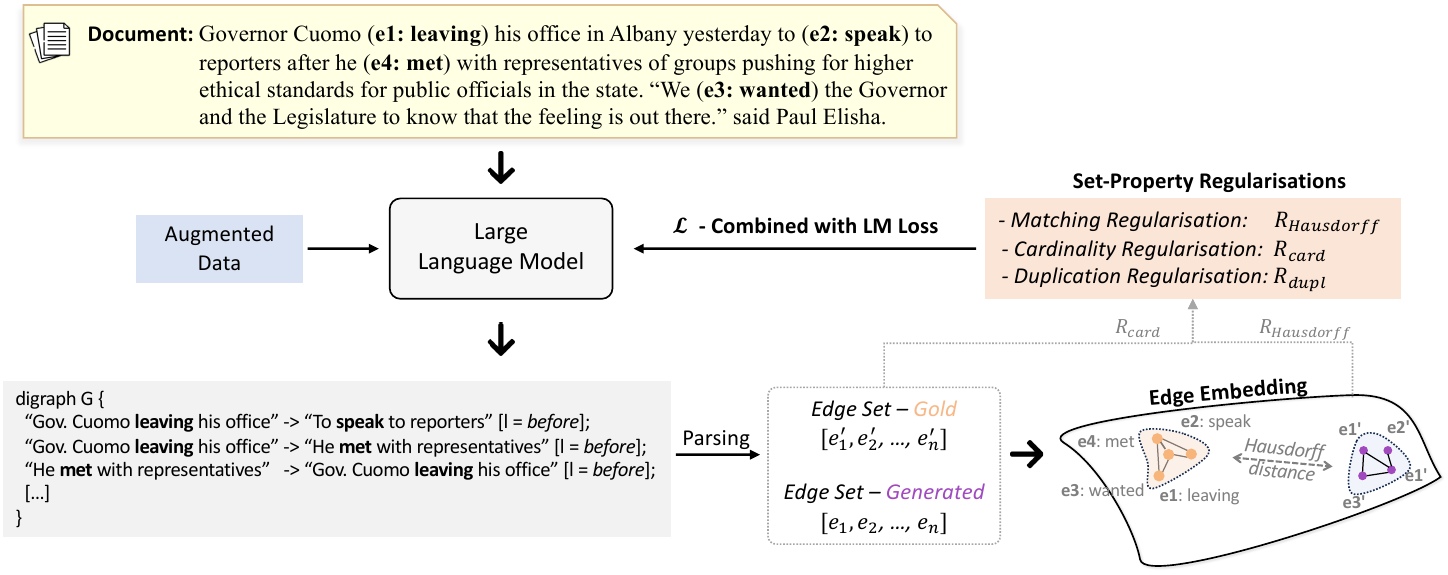}
   \caption{Set-Aligning framework (SAF).} \vspace{-10pt}
   \label{fig:framework}
\end{figure*}

\subsection{Conditional Set Generation}
Text generation models are primarily designed for generating text with strict linear orders, making them suboptimal for generating sets. 
This limitation has been acknowledged in recent NLP research, where efforts have been made to adapt seq2seq frameworks for tasks like multi-label classification and keyword generation \cite{qin-etal-2019-adapting,ye-etal-2021-one2set}.
\citet{DBLP:journals/corr/VinyalsBK15} studied the general challenge of using sets as either input or target output for text generation models.
They found in both cases, the order of elements in the set has a significant impact on convergence and final perplexity.
This implies that there may exist an optimal order for the input or output set sequence, and they proposed allowing the model to search for this order during training.
Instead of resorting to exhaustive search, \citet{madaan-etal-2022-conditional} proposed a data augmentation method to enforce order-invariance and prepend the set's cardinality to the target sequence to ensure the correct cardinality.
While previous research has tackled multi-label prediction and keyphrase generation, our work delves into the unique challenges presented by event temporal graph generation, which involves long sequences and partially ordered properties.

In a more general sense, the object detection task from computer vision also involves set prediction \cite{chen2022pixseq}.
\citet{carion2020end} use parallel decoding to generate the elements in a set based on object queries.
\citet{tan2021sequencetoset} adopted a similar approach in name entity recognition with a non-autoregressive decoder.
Different from the entities in images, the set elements (event relations) in our task are not concrete spacial objects or text spans but instead are varied in length and scattered across each document.
This makes object queries and non-autoregressive decoding inapplicable in our settings.

%%%%%%%%%%%%%%%%%%%%%%%%%%%%%%%%%%%%%%%%%%%%%%%%%%%%%%%%%%%%%%%%%%%%%%%%%%%%%%%
%%%%%%%%%%%%%%%%%%%%%%%%%   Task Definition      %%%%%%%%%%%%%%%%%%%%%%%%%%%%%%
%%%%%%%%%%%%%%%%%%%%%%%%%%%%%%%%%%%%%%%%%%%%%%%%%%%%%%%%%%%%%%%%%%%%%%%%%%%%%%%

%%%%%%%%%%%%%%%%%%%%%%%%%%%%%%%%%%%%%%%%%%%%%%%%%%%%%%%%%%%%%%%%%%%%%%%%%%%%%%%
%%%%%%%%%%%%%%%%%%%%%%%%%   Method        %%%%%%%%%%%%%%%%%%%%%%
%%%%%%%%%%%%%%%%%%%%%%%%%%%%%%%%%%%%%%%%%%%%%%%%%%%%%%%%%%%%%%%%%%%%%%%%%%%%%%%
\section{Set-Aligning Framework}

\citet{madaan-yang-2021-neural} first explored the possibility of end-to-end event temporal graph generation using neural language modelling.
Since then, however, this task has remained under-explored, with numerous unresolved issues.
To elaborate, the first concern is that \citet{madaan-yang-2021-neural} framed graph generation as a conventional sequence generation problem, whereas it is fundamentally a set generation problem.
Secondly, the dataset they built primarily consists of small-sized graphs, failing to challenge the model in terms of document-level understanding.
Lastly, their investigation mainly centred on GPT-2, while the landscape of LLMs has evolved with the emergence of models featuring distinct structures (e.g., encoder-decoder) and new paradigms (e.g., in-context learning) in recent years.
In this study, we address these three aspects to enhance the understanding of sequence-to-sequence temporal graph generation.

Although our proposed framework is designed to be model-independent, several factors have led us to choose Flan-T5 as the base model for our experiments: (i) Based on our preliminary experiments, Flan-T5-base hits the sweet spot in terms of performance vs. resource consumption, allowing us to test more variants; (ii) its encoder-decoder structure is well-suited to document-level graph generation, due to its efficiency in processing comprehensive information in lengthy documents.

\subsection{Event Temporal Graph Modelling as Edge Set Generation}
An event temporal graph is a directed graph with no isolated vertex.
Each edge in the graph describes a temporal relation between two events, and self-loops are not permitted.
Following \citet{madaan-yang-2021-neural}, we represent these graphs by linearizing them into strings using the DOT graph description language~\cite{Gansner2006DrawingGW} (example shown in Figure \ref{fig:framework}). 
Given that event temporal graphs do not have isolated vertices, the sequence essentially represents the edge set of the graph.

We model the probability of generating a string $y$, which is a linearised representation of the event temporal graph $G$, conditioned on a document $X=(x_1,x_2,...,x_n)$ using a language model:
\begin{equation}
    p_{\text{LM}}(y|X)=\prod^{T}_{t=1}{p(y_t|X,y_{<t})}
\end{equation}
\noindent where $y$ is a string formatted in DOT notation.

\subsection{Data Augmentation}
The target sequences of event temporal graph generation are essentially sets rather than strictly ordered text sequences.
Therefore, conventional text generation loss can inadvertently penalise the token order and force the arrangement of elements to match the order in the target sequence, which is not necessarily the optimal order. This enforced order may lead to sub-optimal performance \cite{DBLP:journals/corr/VinyalsBK15}. 
A potential solution is to introduce random permutations of set elements as augmented training examples, which has already been shown effective in tasks like multi-label classification and keyphrase generation  \cite{madaan-etal-2022-conditional}.
Specifically, in the context of event temporal graph generation, the elements correspond to the edges in the target string. The substrings representing the edges are randomly shuffled, while the rest of the string remains unchanged.

Prepending the set cardinality of the ground-truth edge set to the generation target may also help constrain the generation model to avoid over-generation \cite{madaan-etal-2022-conditional}.
However, such attempts in our preliminary experiment led to an approximate $4\%$ drop in edge $F_1$ score, despite a significant reduction in the number of generated edges.
Thus, we decided not to incorporate the cardinality into the final framework.

\subsection{Set Property Regularisations (SPR)}

Simply adding augmented data to train models does not address the fundamental issue of set alignment.
Several challenges arise in this approach. 
First of all, it is unrealistic to add all permutations, especially when dealing with long documents containing numerous event relations, as the training data will grow at a rate proportional to the factorial of the cardinality of the target set.
More importantly, with each augmented example, the loss function would still penalise the unobserved permutations of the set. This would make the training unstable.

The core challenge lies in finding an effective way to compare the linearized target graph with the linearized generated graph, without relying on a strict token-by-token comparison as in conventional text generation.
To tackle this issue, we propose introducing modifications to the generation objective.
As the linearized graph essentially represents the edge set of the graph, we can simplify the graph comparison problem into a set comparison problem. Our approach involves several components. 
Firstly, we add a set cardinality regularisation to encourage the model to generate an adequate number of temporal relation edges. % also mentioned in previous set generation works but is enforce in different way
Then, we introduce a duplication regularisation to penalise any repetition of elements in the edge set.
Lastly, we design a set matching regularisation that assesses the semantic similarity between elements in the target edge set and those in the generated edge set.
Collectively, the above regularisations are referred to as Set Property Regularisations (SPR).
They are integrated with the token-level cross-entropy loss through a weighted average.

To compute the set property regularisations, a graph string needs to be first sampled from a language model given a training input.
Then, this sequence is parsed into a list of edges $E$, where each edge $e$ is a triplet consisting of a head event, a relation type, and a tail event $(h,r,t)$.
The parsing is done with a rule-based parser which turns the graph text string into a structured data representation.
As the edges are loaded in a structured list, the number of edges and duplicated edges can be counted.
Let $\mathcal{E}$ denote the set of all the unique edges in $E$.
The values for the set cardinality regularisation and the duplication regularisation can be computed as follows:

\begin{align}
    \mathcal{E} &= \{e|e\in E\} \\
    R_{\text{dupl}}&=\frac{|E|-|\mathcal{E}|}{|\mathcal{E}|}\\
    R_{\text{card}}&=\frac{\text{abs}(|\mathcal{E}'|-|\mathcal{E}|)}{|\mathcal{E}|}
\end{align}
\noindent where function $\text{abs}(\cdot)$ denotes taking the absolute value, $\mathcal{E}'$ denotes the ground-truth edge set.

To compute the set matching regularisation, we assess the similarity between the generated set and the target set by comparing the semantic similarity of the edges across the two sets.
We take the last layer of the decoder's representations of the respective tokens as the semantic representations of the events and the relation type.
Then, we concatenate these representations as the semantic representation of each edge:
\vspace{-12pt}

\begin{align}
    z_h     & = H_{[h_1,h_2,...,h_m]} \\
    z_r     & = H_{[r_1,r_2,...,r_s]} \\
    z_t     & = H_{[t_1,t_2,...,t_n]} \\
    \bar{e} & = \left[ \text{pool}(z_h); \text{pool}(z_r); \text{pool}(z_t)) \right]
\end{align}
\vspace{-15pt}

\noindent where $H$ is the last-layer hidden states of the decoder. 
$[h_1,...,h_m]$, $[r_1,...,r_s]$, and $[t_1,...,t_n]$ are the indices of the head event, relation type, and tail event, respectively.
$z_h,z_r,z_t$ denote the semantic representations of the head event, relation type, and tail event, respectively.
$\text{pool}(\cdot)$ represents the average pooling function.
$\bar{e}$ denotes the semantic representation of the edge.

We now possess two sets of embeddings: one compassing the edge embeddings extracted from the target graph, and the other containing the edge embeddings derived from the generated graph.
Essentially, they can be considered as two sets of points in the representation space.
Thus, we can measure the similarity of the two graphs by measuring the distance between the two point sets (manifolds) in the representation space.
The Hausdorff distance, originally defined to measure the separation between two subsets within a metric space, has recently found applications in machine learning for measuring the distance between two sets of embeddings \cite{6151115,wang-etal-2023-document}.
We compute the average Hausdorff distance as the measure:

\begin{align}
 d_{H}(\mathcal{E}', \mathcal{E}) =& \frac{1}{|\mathcal{E}'|}\sum_{\bar{e'}\in\mathcal{E}'}\min_{\bar{e}\in\mathcal{E}}d_{cos}(\bar{e'},\bar{e}) \nonumber \\ 
 &+ \frac{1}{|\mathcal{E}|}\sum_{\bar{e}\in\mathcal{E}}\min_{\bar{e'}\in\mathcal{E}'}d_{cos}(\bar{e'},\bar{e})
\end{align}

\noindent where the distance of an edge pair is computed by the cosine distance $d_{cos}(\cdot)$.

When the model generates the set elements in a different order than the target sequence, the token-level cross-entropy loss would be high.
If the model generates more correct elements as suffixes of the sequence with a wrong order, the loss value would probably increase further.
However, the SPR will have a lower value and thus alleviate the discouragement for generating more elements caused by the token-level cross-entropy loss.

\subsection{Fine-tuning with Set Property Regularisations}

Unlike the set prediction methods based on parallel decoding \cite{carion2020end,tan2021sequencetoset}, SPR cannot be directly used as the main objective in auto-regressive generation.
There are two primary reasons for this. 
The first reason is that obtaining the SPR requires sampling from the decoder, which would reduce the training speed significantly.
Moreover, the second reason is that the language model will struggle to generate sequences in DOT format accurately because learning the token dependency for such format requires the language modelling objective.
Consequently, the sequence parser will fail to recognize any valid edges within the sequence, resulting in high SPR values and hindering the training. 

To avoid the problems mentioned above, we introduce the SPR after a certain number of fine-tuning iterations.
Once the model has acquired a basic proficiency in generating correct DOT sequences, the SPR can function as intended. 
SPR can prevent the language model from overfitting to the order of the target set shown in the training samples.

We explored alternative approaches to incorporate SPR, but they reported inferior performance compared to the method eventually included in our framework.
We discuss those alternative methods in the Appendix \ref{alternatives}.

%%%%%%%%%%%%%%%%%%%%%%%%%%%%%%%%%%%%%%%%%%%%%%%%%%%%%%%%%%%%%%%%%%%%%%%%%%%%%%%
%%%%%%%%%%%%%%%%%%%%%%%%%   Experiment and Result        %%%%%%%%%%%%%%%%%%%%%%
%%%%%%%%%%%%%%%%%%%%%%%%%%%%%%%%%%%%%%%%%%%%%%%%%%%%%%%%%%%%%%%%%%%%%%%%%%%%%%%
\section{Experiment}

\subsection{NYT Temporal Event Graph Dataset}

\begin{table}[htb]
  \begin{center}
  \resizebox{\columnwidth}{!}{%{\small
  \begin{tabular}{lrrr}
  \toprule
  \bf ~ & NYT-train & NYT-test & NYT-human \\ 
  \midrule 
 Total documents & $18,263 $  & $1,000$ & $22$\\
 Total events & $846,022 $ & $47,251$ & $661$\\
 Node degree & $2.52$ & $2.54$ & $2.34$ \\
 Total relations & $1,066,264 $ & $60,056$ & $528$\\
 \textit{before} & $578,216$ & $32,729$ & $465$ \\
 \textit{after} & $412,704$ & $23,200$ & $0$ \\
 \textit{includes} & $7,922$ & $450$ & $12$ \\
 \textit{is\textunderscore included} & $41,964$ & $2,332$ & $0$\\
 \textit{simultaneous} & $25,458$ & $1,345$ & $51$\\
  \bottomrule
  \end{tabular}}
  \end{center}
  \caption{The statistics of the NYT temporal event graph dataset. Node degree represent the average number of relations each event has. }% \vspace{-12pt}}
  \label{data_table}
\end{table}

There are several event temporal relation extraction datasets with pairwise event relation annotations, such as MATRES and TBD.
It is theoretically possible to convert these annotations into document-level event temporal graphs.
However, our preliminary experiments have shown that even when merging all of these datasets (resulting in 4,684 training documents), it is not sufficient to fine-tune a large language model to achieve acceptable performance.
To address this limitation, we opted to build a significantly larger dataset on a selection of data from the New York Times (NYT) corpus using a weak supervision approach, drawing inspiration from the work of \citet{madaan-yang-2021-neural}.
Nevertheless, we introduced additional steps in the data selection process to ensure that the selected documents contain high-quality event temporal graphs, which were not taken in \citet{madaan-yang-2021-neural}.

Firstly, we performed topic modelling using Latent Dirichlet Allocation (LDA) on the MATRES and TBD datasets to extract a set of topics. 
Then, we identified general descriptors that are semantically similar to these topics (e.g., politics, diploma, sports, etc.). This selection process was crucial because, following training with noisy labels, our intention was to evaluate the model's performance on these datasets under zero-shot settings.
We further analysed the most noteworthy events in these descriptors to ensure they were narrative-oriented, because articles that weave stories tend to contain a wealth of event temporal relations.
To identify the most significant events, we employed a metric similar to TF$\cdot$IDF which we could describe as ``event frequency $\times$ inverse-descriptor frequency''. 

\begin{equation}
    \text{ef}\cdot \text{idf}=\frac{f_{\mathfrak{e},d}}{\sum_{\mathfrak{e}'\in d}f_{\mathfrak{e}',d}}\cdot \log \frac{|D|}{|\{d\in D:\mathfrak{e}\in d\}|}
\end{equation}
\noindent where $\mathfrak{e}$ is an event and $d$ is a descriptor. 
$f_{\mathfrak{e},d}$ is the number of times that event $\mathfrak{e}$ occurs in the documents with the descriptor $d$.
$\sum_{\mathfrak{e}'\in d}f_{\mathfrak{e}',d}$ is the total number of event occurrence in the descriptor $d$.
$|D|$ is the total number of descriptors in the corpus.
$|\{d\in D:\mathfrak{e}\in d\}|$ is the number of descriptors where the event $\mathfrak{e}$ appears.

The descriptors that are selected and the number of documents in them are listed in the Appendix \ref{sec:appendix_1}. % add in appendix
After choosing the documents, we acquire the event temporal graph by running an off-the-shelf event and temporal relation extraction tool called CAEVO \cite{mcdowell-etal-2017-event}.
CAEVO is more scalable than Cogcomptime \cite{ning-etal-2018-cogcomptime}, making it suitable for building a large-scale dataset.

Then, each temporal graph is represented in DOT format, and every event verb is prefixed and suffixed with its noun phrase and object, respectively.
Note that we did not break the documents into short segments as \citet{madaan-yang-2021-neural} did.
Instead, we keep the data strictly at the document level which is a more challenging setting because the model needs to analyse the entire document and generate a much larger graph.
In the dataset we built, a target graph has about $46$ nodes and $58$ edges on average.
While in \citet{madaan-yang-2021-neural}, the average number of nodes is $4$ and the average number of edges is $5$ in a document-level event temporal graph.
Moreover, their events have $1.54$ relations on average, while events in our data have $2.52$ relations on average, showing that the graphs in our dataset are much more complex.
In practice, these complex documents are the ones that require analysis, and a model developed based on simpler inputs cannot handle them directly.

\subsection{Human-annotated Test Data}
Aside from testing with the CAEVO-created data, we recruited human annotators to annotate a test split of the NYT data.
We performed a preprocessing step regarding the relation types by merging the reciprocal relations, such as transforming \textit{after} into \textit{before}, \textit{is\textunderscore included} into \textit{includes} by swapping the head and tail events.
For example, ``I had dinner after I had lunch'' is equivalent to ``I had lunch before I had dinner''.
This processing not only streamlined the annotation process but also enhanced the model performance (refer to experimental results in Appendix \ref{additional_results}).
We recruited crowd workers from Prolific\footnote{\href{http://www.prolific.com}{prolific.com}} platform, which is a research-focused platform providing verified human workers.
We recruited $24$ participants in total (including pilot testing runs).
To make sure the participants can understand and annotate the article efficiently, we only recruited native English speakers who have an education level higher than High school diploma/A-levels.
We put $4$ documents, which are randomly sampled from the same descriptor set as the training and testing of the selected NYT corpus, into each unit task.
There is a shared document across all the tasks to compute the inter-annotator agreement (IAA).
To minimize discrepancy, we asked $2$ participants to first identify the event triggers in each unit task.
We then merged the event annotations from the participants by taking the union of the spans (if there are overlapped spans, we take the longer span).
Then, we asked another participant to annotate the event temporal relation based on the identified events.
We also included the outputs from the CAEVO model to serve as examples, but we explicitly asked the participants to correct the annotations by adding, removing, or changing the CAEVO's annotations.
In the end, we collected $22$ documents as the human-annotated test set.
On the event identification, we compute IOU (Intersection over Union) as a measure of agreement between the annotators.
Average across $7$ tasks, the IOU between the event spans is $0.8986$.
For the relation annotations, we compute the average Cohen's $\kappa$ of every participant pair in the relation annotation task (on the shared document).
The average Cohen's $\kappa$ is $0.7465$.
Details of instructions and interfaces are in Appendix \ref{sec:appendix_1}.

The statistics of the constructed datasets are shown in Table \ref{data_table}.
The distributions of relation types are highly imbalanced, with a majority falling into either the \textit{before} or \textit{after} categories.
We also evaluated the trained models on the MATRES test set (comprising $20$ documents) and TBD test set (consisting of $9$ documents), both of which are based on human annotations and processed into DOT using the methods previously described.

\subsection{Model Setting}
We employed \textbf{Flan-T5-base} as the backbone model for contextualised graph generation.
We first trained a Flan-T5-base model following the same setup as in \citet{madaan-yang-2021-neural} as the baseline.
\textbf{SAF (w/o DA)} is our proposed framework without the augmentations of edge order but with Set Property Regularisations (SPR).
\textbf{SAF (w/o SPR)} is the framework without the use of SPR but with the augmentations. 
As our SAF framework with SPR requires additional training steps and the augmentations enlarge the training set, we keep the number of training steps balanced in the methods to exclude the influence of seeing different amounts of training data.
The model is trained for $10$ epochs, with each document being augmented through $4$ random permutations, followed by a further $3$ epochs of training, during which the SPR are adopted without permutations.
We use a learning rate of $2e-5$, along with a weight decay of $0.01$.
Batch size of $5$ before SPR, and $3$ during SPR because additional memory is required for sampling.
We used AdamW optimizer \cite{loshchilov2018decoupled}.
We use the beam search~\cite{Graves2012SequenceTW} with a beam size of $5$ and a maximum length of $2048$ to sample results.
We balanced the training steps in the compared methods to make sure they saw the same amount of training data.
Experiments are conducted on a GPU node under an HPC cluster using $4$ Nvidia A100 80G GPUs.
The models are trained based on $3$ random seeds (ChatGPT was tested for $3$ times) and the metrics are the average values of them.
Training with the augmented data for $10$ epochs requires approximately $19$ hours. 
Training with SPR for $3$ epochs takes about $27$ hours. 
Training a vanilla Flan-T5-base for the same number of training steps demands approximately $20$ hours.

\subsection{Evaluation Metrics}
Following the previous research~\cite{madaan-yang-2021-neural}, we evaluate the results using the metrics of precision, recall, and $F_1$ score for both node set and edge set predictions.
The primary metric is the edge $F_1$ because the quality of the node generation is also reflected in it.

\begin{table}[tb]
  \begin{center}
  \resizebox{\columnwidth}{!}{%{\small
  \begin{tabular}{lrrrrrr}
  \toprule
  & \multicolumn{3}{c}{NYT-test}  & \multicolumn{3}{c}{NYT-human}\\
  \cmidrule(lr){2-4} \cmidrule(lr){5-7} 
  \bf ~ & $P^E$ & $R^E$ & $F^E_1$ &$P^E$ & $R^E$ & $F^E_1$\\ 
  \midrule 
Flan-T5-base & $51.27$& $32.43$  & $39.73$ &$22.61$ & $25.88$& $24.14$\\
  SAF (w/o DA) & $50.28$& $34.82$ & $41.15$& $25.80$&$32.13$& $28.62$\\
  SAF (w/o SPR) & $\bf{51.88}$& $36.64$& $42.95$ &$\bf{27.08}$&$34.91$&$30.50$\\
  SAF & $50.97$& $\bf{39.96}$  & $\bf{44.80}$ & $25.92$&$\bf{40.21}$&$\bf{31.52}$\\
  \bottomrule
  \end{tabular}}
  \end{center}
  \caption{Edge-based metrics on the NYT datasets}% \vspace{-12pt}}
  \label{NYT_dist_table}
\end{table}

\begin{table}[tb]
  \begin{center}
  \resizebox{\columnwidth}{!}{%{\small
  \begin{tabular}{lrrrrrr}
  \toprule
  & \multicolumn{3}{c}{NYT-test}  & \multicolumn{3}{c}{NYT-human}\\
  \cmidrule(lr){2-4} \cmidrule(lr){5-7} 
  \bf ~ & $P^N$ & $R^N$ & $F^N_1$ &$P^N$ & $R^N$ & $F^N_1$\\ 
  \midrule 
Flan-T5-base     & $\bf{75.52}$ & $58.24$  & $65.76$&$53.36$&$47.66$&$50.35$\\
  SAF (w/o DA) & $75.34$ & $60.64$  & $67.20$     &$\bf{54.86}$&$50.43$&$52.55$ \\
  SAF (w/o SPR) & $75.43$ & $62.36$  & $68.27$     &$54.14$&$51.59$& $52.84$\\
  SAF           & $75.47$&$\bf{65.16}$&$\bf{69.95}$&$53.63$&$\bf{54.51}$&$\bf{54.06}$\\
  \bottomrule
  \end{tabular}}
  \end{center}
  \caption{Node-based metrics on the NYT datasets}% \vspace{-12pt}}
  \label{NYT_dist_table_node}
\end{table}
\subsection{Results}
\begin{figure}[tb]
    \centering
    \includegraphics[width=\columnwidth]{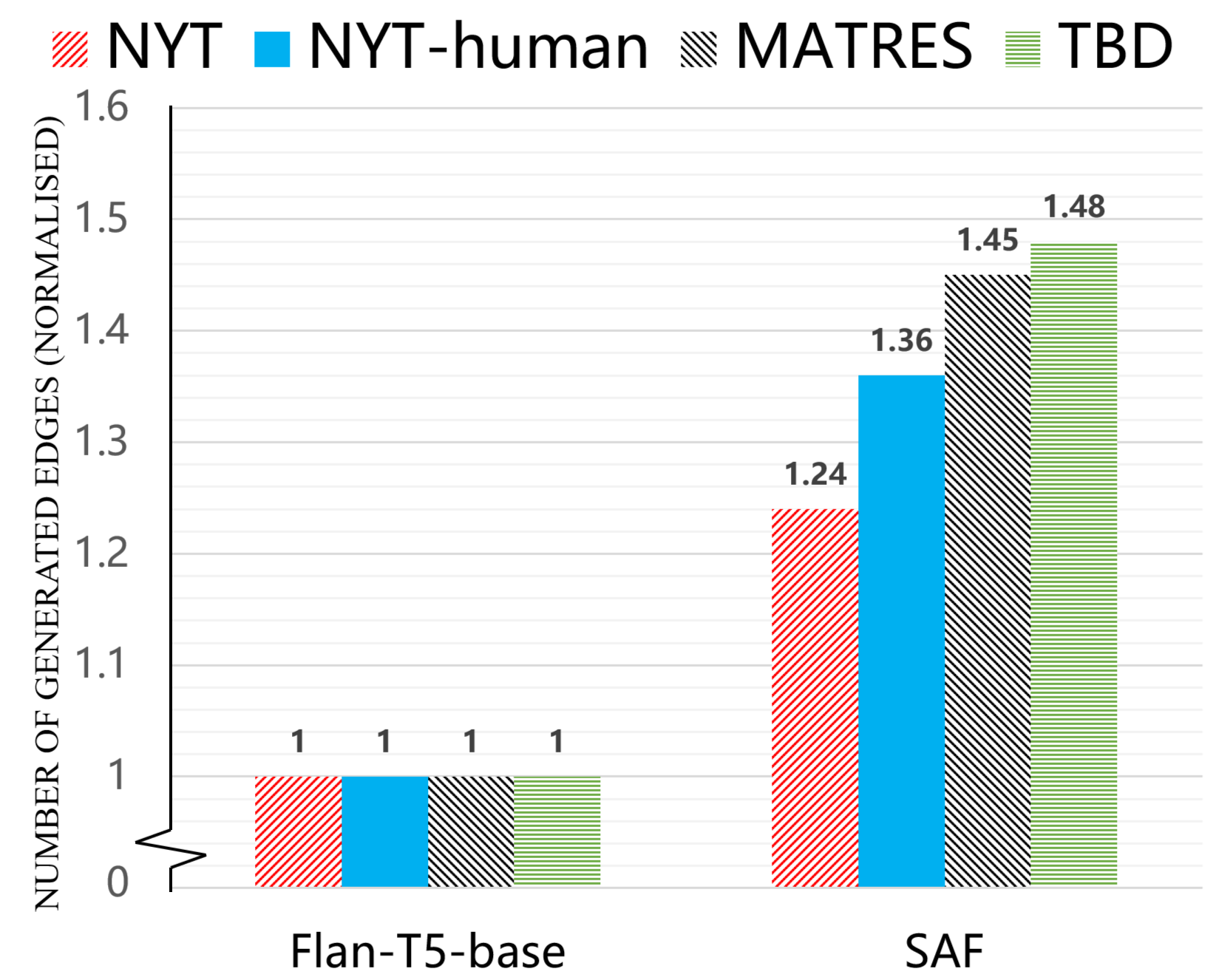}
    \caption{The comparison of generated edges between SAF and vanilla Flan-T5-base. The $y$ axis is normalised by dividing the number of edges generated by Flan-T5-base in the respective datasets.}
    \label{fig:edges}
\end{figure}

As shown in table \ref{NYT_dist_table} and \ref{NYT_dist_table_node}, SAF (w/o SPR) consistently outperform Flan-T5-base in terms of $F_1$ scores on the NYT-test and NYT-human datasets, suggesting the benefits of introducing permutated training examples.
For example, SAF (w/o SPR) improves upon Flan-T5-base by about $3\%$ on the NYT-test and $6\%$ on the NYT-human in terms of edge $F_1$.
SAF (w/o DA) achieves an improvement of approximately $1.5\%$ on the NYT-test and $4.5\%$ on the NYT-human datasets in terms of edge $F_1$, demonstrating the effectiveness of SPR alone. 
Furthermore, our SAF model yields the best performance when both SPR and augmentation are incorporated.
We also observe that models utilizing SAF have much higher edge recalls while their edge precision scores are either similar or occasionally even lower than those of other models. This suggests that the performance improvement primarily comes from the generation of more edges.
This observation is reinforced by the information presented in Figure \ref{fig:edges}, where models trained with SAF can generate $24\%-48\%$ more edges compared to the conventional text generation framework on these datasets.
These additional edges play a pivotal role in the improvement of the edge $F_1$ since precision stays nearly the same.

It is worth mentioning that the NYT-human dataset has a different label distribution compared to the NYT dataset used for training, where its events and event temporal relations were produced by CAEVO.
Notably, the frequency of \textit{simultaneous} is significantly higher, accounting for $9.66\%$, in contrast to the $2.39\%$ observed in the training set (see Appendix \ref{sec:appendix_1} for more comprehensive analyses). 
Based on our observation, it appears that human annotators tend to apply a more lenient criterion for the \textit{simultaneous} label whereas CAEVO enforces a stricter definition of this label.

\begin{table}[tb]
  \begin{center}
  \resizebox{\columnwidth}{!}{%{\small
  \begin{tabular}{lrrrrrr}
  \toprule
  & \multicolumn{3}{c}{MATRES-test}  & \multicolumn{3}{c}{TBD-test}\\
  \cmidrule(lr){2-4} \cmidrule(lr){5-7} 
  \bf ~ & $P^E$ & $R^E$ & $F^E_1$ &$P^E$ & $R^E$ & $F^E_1$\\ 
  \midrule 
ChatGPT& $10.58$ & $6.56$ & $8.09$ &$25.92$ &$5.94$  & $9.66$ \\
    Flan-T5-base & $13.06$ & $7.16$ & $9.25$ & $23.26$ & $4.59$ & $7.67$\\
    SAF & $\bf{18.05}$ & $\bf{14.31}$ & $\bf{15.96}$ & $\bf{37.53}$ &$\bf{11.04}$& $\bf{17.05}$\\
  \bottomrule
  \end{tabular}}
  \end{center}
  \caption{Experiment results on human-annotated MATRES and TBD under the zero-shot setting. }
  \label{matres_table}
\end{table}

Similar trends are also observed in Table \ref{matres_table}, which were obtained through evaluation on MATRES and TBD. 
We used the models trained on the NYT training set to test on these datasets under zero-shot settings.
ChatGPT shows our best attempts to generate event temporal graphs with gpt-3.5-turbo model through Openai API. 
We used two hops: (i) ask ChatGPT to generate events from the documents, (ii) ask ChatGPT to generate an event temporal graph based on the generated events and the documents.
The results show that ChatGPT is outperformed by fine-tuned models, which is in line with the recent papers on exploring ChatGPT's ability on event understanding \cite{li2023evaluating,chan2023chatgpt,gao2023exploring}.

Upon examining the responses of ChatGPT, it appears that it conceptualises events as a broader and high-level notion which diverges from the definition commonly used by the information extraction community in event processing.  
In our task, each predicate can signify an event, but ChatGPT tends to approach event identification more like a summarisation task, where it summarises text chunks in a document. 
This likely explains why ChatGPT identified only approximately half of the events present in the target graph, resulting in low recall. 
These observations confirm that event temporal graph generation cannot be solved solely with prompt engineering on ChatGPT.
It is noteworthy that ChatGPT consistently produces graphs in the correct DOT format across both the MATRES-test and TBD-test datasets, indicating that formatting issues are not the primary factor in ChatGPT's underperformance. 
The details regarding the inputs, outputs, and parameter settings for ChatGPT are presented in Appendix \ref{chatgpt_discussion}.

\subsection{Error Analysis}
A major error type we found is that the model often fails to deduce temporal relationships that involve inference.
This is due to the reliance of weak supervision signals provided by CAEVO, which primarily rely on syntactical rules.
Consequently, this problem led to a lower edge $F_1$ on the human-annotated test set, as human annotators provided many temporal relations that were inferred through commonsense reasoning.
For example, the model does not perceive a clear temporal sequence in the sentence:``<person A> won the gold medal in women's 1,500m. <person B> won the silver and <person C> won the bronze.''
However, human annotators can readily identify an obvious temporal order among ``<person A> won'', ``<person B> won'', and ``<person C> won'', as it aligns with the common knowledge that in a race, the first person who crossed the finish line won the gold, followed by the silver and the bronze winners.
% refer to Appendix
% a figure or a schematic graph

\section{Conclusion}

This study proposes a framework for fine-tuning language models to generate event temporal graphs directly from raw documents in an end-to-end manner.
The proposed framework includes a data augmentation method and set property regularisations to mitigate the problem caused by conventional generation loss, promoting the generation of more edges by language models and, consequently, leading to improved performance. 
Extensive experiments show the effectiveness of our proposed model on multiple widely used datasets with real-world articles.
The thorough analysis demonstrates that our framework can encourage language models to generate more edges for constructing event temporal graphs in various settings.

\section*{Limitations}
Due to the presence of noisy labels used in fine-tuning, a major limitation of the proposed method is the inclusion of many imaginary events, trivial events, and negative expressions of events.
For example, CAEVO identified phrases like ``<someone> did not \textbf{fire}'' as an event.
While ``fire'' serves as a predicate and the notion of ``did not \textbf{fire}'' can hold narrative significance, it may not be entirely suitable within the context of event temporal graphs.
This is because it is not about the occurrence of an action or a change of state, but rather describes the absence of an event.
Similarly, in some articles, there are descriptions of multiple potential future developments, such as "he might \textbf{buy} product A".
Including such expressions as events might introduce confusion into the event temporal graph, as these represent possibilities rather than actual occurrences.
This problem mainly arises from the behaviour of the CAEVO method,  which primarily focuses on identifying fine-grained predicates as events.
The resolution to this problem lies in obtaining better-quality supervision signals which focus on salient events (i.e., events which are mentioned frequently and are important to the narrative).

\section*{Ethics Statement}
The proposed method analyses the text provided and extracts relevant information from it.
The algorithm cannot acquire information beyond the boundary of the given text.
Thus, any associated risks stem solely from the data itself.
This research only utilised publicly available data.
As long as the data input to the model is collected according to the relevant data policies and guidelines, the proposed method does not introduce further risks.

\section*{Acknowledgements}

This work was supported in part by the UK Engineering and Physical Sciences Research Council through a Turing AI Fellowship (grant no. EP/V020579/1, EP/V020579/2). 
Computing facilities were provided by the Scientific Computing Research Technology Platform of the University of Warwick, and the CREATE HPC platform of the King's College London \cite{Kings2024CREATE}.

\bibliography{anthology,custom}
\bibliographystyle{acl_natbib}

\newpage
\clearpage

\appendix

\setcounter{table}{0}
\renewcommand{\thetable}{A\arabic{table}}

\setcounter{figure}{0}
\renewcommand{\thefigure}{A\arabic{figure}}

\section{Discussion of Alternative Approaches for Incorporating Set Property Regularisation (SPR)}
\label{alternatives}

In our preliminary studies, we tested several way of incorporating SPR into the training process.
We tried using the weighted average of SPR and the language modelling loss in every training step (the weight of SPR increases across the training process). 
However, the training becomes very slow due to the added decoding processes.

We also experimented with introducing randomness into the incorporation of the SPR in some of the training steps. 
In each training step, there was a $0.5$ probability that the SPR were computed and backpropagated, while in other cases, only the language modelling loss was considered. 
This probability increased progressively with the epoch number. 
For example, during the initial epoch (epoch $1$), the probability was set to 0, as we explained in Section 3.4, the model struggled to generate outputs in the correct format during the early stages. 
Subsequently, the probability was increased linearly, reaching $1$ by the last epoch.

However, this approach proved to be ineffective and caused training instability.
The loss value fluctuated between steps, leading to confusion for the model. 
It is worth noting that this was an initial experiment and significantly differed from the final version of the proposed method.

\section{Additional Error Analysis}
We observed a type of error involving the model's incorrect prediction of long-distance temporal relationships.
The model sometimes predicts a temporal relation between two events that are separated by more than ten sentences.
This is unexpected, as the CAEVO model, which produces weak supervision signals, typically does not extract relations for events that are more than two sentences apart from each other. In essence, it primarily focuses on events within close proximity.
%It only extracts temporal relations from the events that are within two sentences away from each other.
Our observations suggest that human annotators also tend not to annotate temporal relations for events that are distant from each other, arguably because such relations are often implicit and can be challenging to track across large chunks of text.
% worth putting more details about this?

% analysis of the constraint violation

\section{More Analysis about the Generation Results}
\label{additional_results}
Table \ref{reciprocal_table} shows a preliminary experiment in which the augmented Flan-T5-base models were trained and tested on NYT-test before and after merging reciprocal relation types.
The first row is the model trained with \textit{before}, \textit{after}, \textit{includes}, \textit{is\textunderscore included}, and \textit{simultaneous}.
The second row is the model trained by merging \textit{after} with \textit{before}, \textit{is\textunderscore included} and \textit{includes} by swapping the head and tail events.
Both models are trained with $4$ augmented instances for each original instance.
The results show the model benefits from the simpler label set.

\begin{table}[h]\small
  \begin{center}
  \resizebox{\columnwidth}{!}{%{\small
  \begin{tabular}{lrrrrrr}
  \toprule
  \bf ~ & $P^N$ & $R^N$ & $F^N_1$ & $P^E$ & $R^E$ & $F^E_1$\\ 
  \midrule 
  With reciprocal relations & $76.05$ & $55.05$ & $63.87$ & $52.07$& $30.90$  & $38.78$\\
  Merge reciprocal & $75.48$ & $61.96$  & $\bf{68.05}$ & $52.03$& $36.34$  & $\bf{42.79}$\\
  \bottomrule
  \end{tabular}}
  \end{center}
  \caption{Comparison between model trained with reciprocal relations or merging reciprocal relations.}% \vspace{-12pt}}
  \label{reciprocal_table}
\end{table}

Table \ref{distribution_table} shows the relation type distribution generated by the models.
The relation distributions are highly unbalanced.

\begin{table}[h]\small
  \begin{center}
  \resizebox{\columnwidth}{!}{%{\small
  \begin{tabular}{lrrr}
  \toprule
  \bf ~ & \textit{before} & \textit{includes} & \textit{simultaneous} \\ 
  \midrule 
  Target graph & $93.13$ & $4.63$ & $2.24$\\
  Flan-T5-base & $92.82$ & $3.19$ & $3.99$\\
  T5-base & $92.48$ & $3.94$ & $3.59$\\
 SAF (Flan-T5-base) & $93.45$ & $3.28$ & $3.27$\\
 SAF (T5-base) & $93.00$ & $3.65$ & $3.35$\\
  \bottomrule
  \end{tabular}}
  \end{center}
  \caption{Generated graph temporal relation label distribution (in percentage).}% \vspace{-12pt}}
  \label{distribution_table}
\end{table}

Table \ref{degree_table} shows the average degree for the nodes in the generated graphs.
SAF generates more complex graphs with a higher average node degree than the compared approaches.

\begin{table}[h]\small
  \begin{center}
  \resizebox{\columnwidth}{!}{%{\small
  \begin{tabular}{lr}
  \toprule
  \bf ~ & average node degree \\ 
  \midrule 
  Flan-T5-base & $2.06$\\
SAF (w/o SPR) & $2.16$\\
 SAF & $2.31$\\
  \bottomrule
  \end{tabular}}
  \end{center}
  \caption{The average node degree of the generated graphs on NYT.}% \vspace{-12pt}}
  \label{degree_table}
\end{table}

We also investigate the effect of the order of the target edge set (Table \ref{NYT_order}).
The first row is Flan-T5-base fine-tuned with the target edge set ordered based on random order in the documents.
The second row is Flan-T5-base fine-tuned with target edge set ordered based on their appearance order in the documents.
We could observe that the appearance order results in slightly better performance than the random order.
Each method was run on $5$ different random seeds and trained for $50$ epochs.
\begin{table}[h]
  \begin{center}
  \resizebox{\columnwidth}{!}{%{\small
  \begin{tabular}{lrrrrrr}
  \toprule
  & \multicolumn{3}{c}{NYT-test}  & \multicolumn{3}{c}{NYT-human}\\
  \cmidrule(lr){2-4} \cmidrule(lr){5-7} 
  \bf ~ & $P^E$ & $R^E$ & $F^E_1$ &$P^E$ & $R^E$ & $F^E_1$\\ 
  \midrule 

  Random order & $51.02$& $29.37$ & $37.28$& $18.15$&$20.45$& $19.24$\\
  Appearance order & $51.20$& $30.24$  & $38.02$ &$19.24$ & $21.78$& $20.43$\\
  \bottomrule
  \end{tabular}}
  \end{center}
  \caption{Comparison based on different sequence orders on NYT-test and NYT-human.}% \vspace{-12pt}}
  \label{NYT_order}
\end{table}

\section{Annotation of the Test Set}
\subsection{Overview}
\label{sec:appendix_1}
We recruited crowd workers from Prolific\footnote{\href{http://www.prolific.com}{prolific.com}} platform, which is a research-focused platform providing verified human workers.
We recruited $24$ participants in total (including pilot testing runs).
In order to make sure the participants can understand and annotate the article efficiently, we require the participants to be native English speakers and have an education level higher than High school diploma/A-levels.
We put $4$ documents, which are randomly sampled from the same descriptor set as the training and testing of the selected NYT corpus, into each unit task.
There is a shared document across all the tasks for the purpose of computing the inter-annotator agreement (IAA).
In order to maximize the IAA, we asked $2$ participants to first identify the event triggers in each unit task.
After that, we merged the event annotations from the participants by taking the union of the spans (if there are overlapped spans, we take the longer span).
Then, we asked another participant to annotate the event temporal relation based on the identified events.
We also included the outputs from the CAEVO model to serve as examples, but we explicitly asked the participants to correct the annotations by adding, removing, or changing the CAEVO's annotations.
In the end, we collected $22$ documents as the human-annotated test set.

On the event identification, we compute IOU (Intersection over Union) as a measure of agreement between the annotators.
Average across $7$ tasks, the IOU between the event spans is $0.8986$.
For the relation annotations, we compute the average Cohen's $\kappa$ of every participant pair in the relation annotation task (on the shared document).
The average Cohen's $\kappa$ is $0.7465$.

\subsection{Chosen Descriptors}
Here are the chosen descriptors: ``airlines and airplanes'', ``olympic games'', 
                      ``tennis'', ``united states international relations'',
                      ``international relations'', ``civil war and guerrilla warfare'',
                      ``track and field'', ``soccer'', ``bombs and explosives'',
                      ``politics and government''.
We choose $2,000$ documents from each descriptor.
After preprocessing and filtering out some invalid documents, we have $18,263$ documents in NYT-train, $1,000$ documents in NYT-test, and $22$ documents in NYT-human.

\subsection{Instructions and Interface}
\begin{figure*}[h] % 'h' for here
   \centering
   \includegraphics[width=0.95\textwidth]{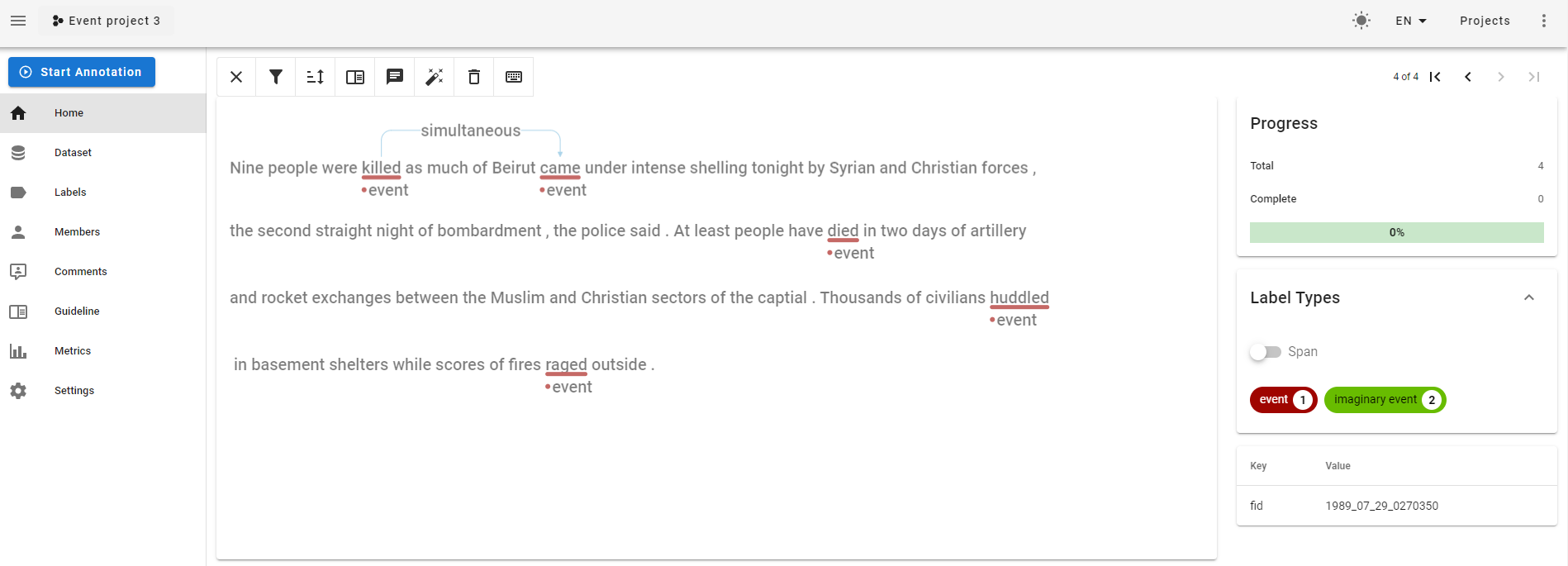}
   \caption{Annotation interface for event identification.}
   \label{fig:event}
\end{figure*}

We use a popular open-sourced annotation interface called Doccano.
As shown in Figure \ref{fig:event}, annotators can select text spans for events.
To direct annotators to distinguish events that actually occurred and imaginary events, we also provide an ``imaginary event'' label type.
We asked them to annotate the predicates that are about a negative expression of an action or just a hypothesis in the context as an imaginary event.
Imaginary events are orthogonal to the real-world timeline and thus have limited meaning for understanding the narrative.

\begin{figure*}[h] % 'h' for here
   \centering
   \includegraphics[width=0.95\textwidth]{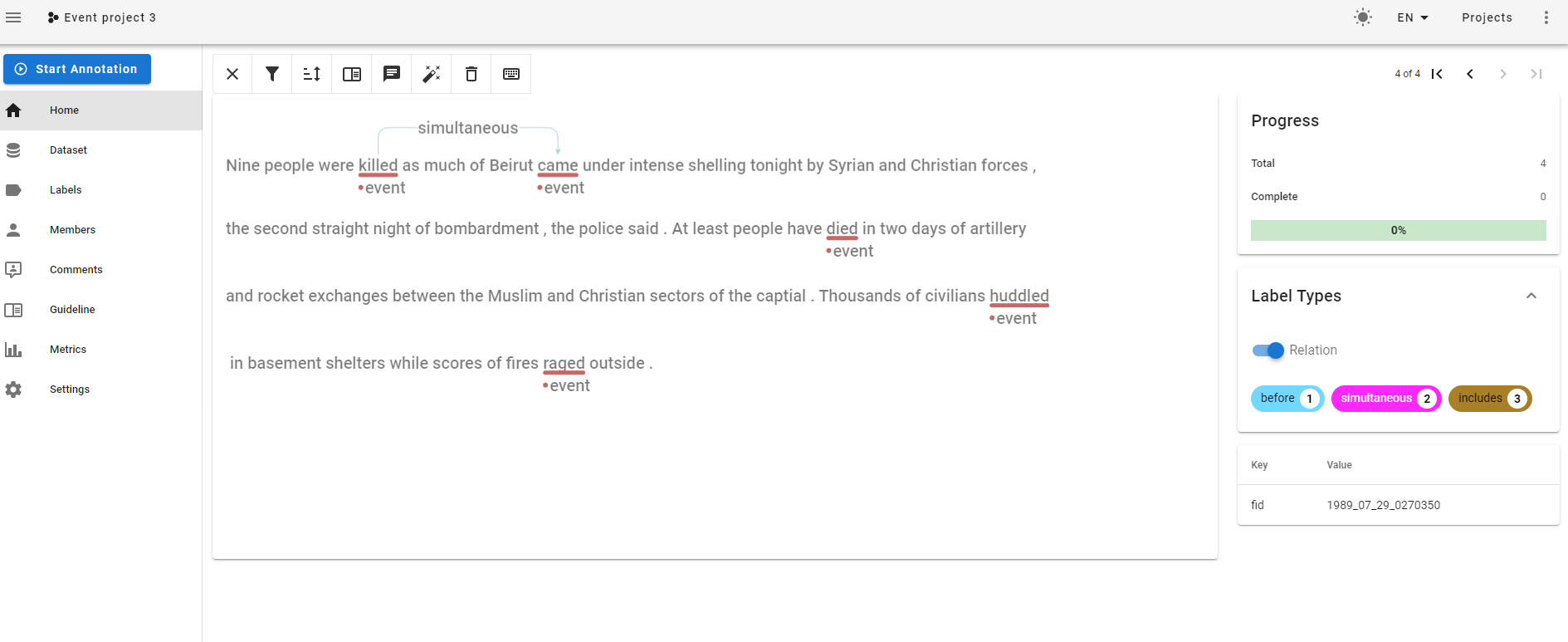}
   \caption{Annotation interface for event relation identification.}
   \label{fig:relation}
\end{figure*}

Figure \ref{fig:relation} shows the interface for annotating the relation.
On this page, annotators can select two existing event spans, and then select the relation type from ``before'', ``includes'', and ``simultaneous''.

Before the annotators came to the annotation platform, they went through a website where we put detailed descriptions and terminology definitions about the task.
We also provided a video tutorial for using the annotation platform.

\section{ChatGPT prompting}
\label{chatgpt_discussion}
We used the OpenAI API chat completion model \emph{gpt-3.5-turbo-0613}.
We used the ``function call'' method to ensure better parsing quality.
The function call parameters are shown in Figure \ref{fig:function_call}.
The temperature is set to $0$.
The other parameters are set as default.
We show the inputs and outputs of the multi-hop prompting in Table \ref{prompt_table}.

\section{GPT-4 Case Study}

We present various test cases of prompting with GPT-4 through this link\footnote{\href{https://chat.openai.com/share/1dd5ede4-f7b1-4b3c-8c91-e02cee15c523}{Test cases on the TBD dataset.}}.
The responses from GPT-4 essentially serve as summaries of the documents provided.
The events it understood are quite broad, akin to abstracts of segments in the documents.
This diverges from the NLP community's definition of event understanding, which typically focuses on specific action occurrences and aims to obtain more granular information within the event temporal graph.

Another notable aspect of the graphs generated by GPT-4 is its tendency to represent a linear sequence of items ordered by their appearances in the document.
This ties back to the first issue concerning how GPT-4 comprehends events.
It essentially generated a summary of the document, which, while not incorrect, does not adhere to the standard of event temporal graph extraction defined in SemEval 2013 TempEval-3\cite{uzzaman-etal-2013-semeval}.

Simply providing the definition of an event has not resulted in a change in its behaviour\footnote{\href{https://chat.openai.com/share/38433651-80b3-4d3d-bba0-0bf5dc8ec596}{Prompting with event definition.}}.
While extensive prompt engineering might help, we believe that incorporating certain supervision signals could still be necessary.
Our framework could prove valuable for instruction finetuning, aligning specific instructions with the event temporal graph generation task.

\begin{figure*}[h] % 'h' for here
   \centering
   \includegraphics[width=0.95\textwidth]{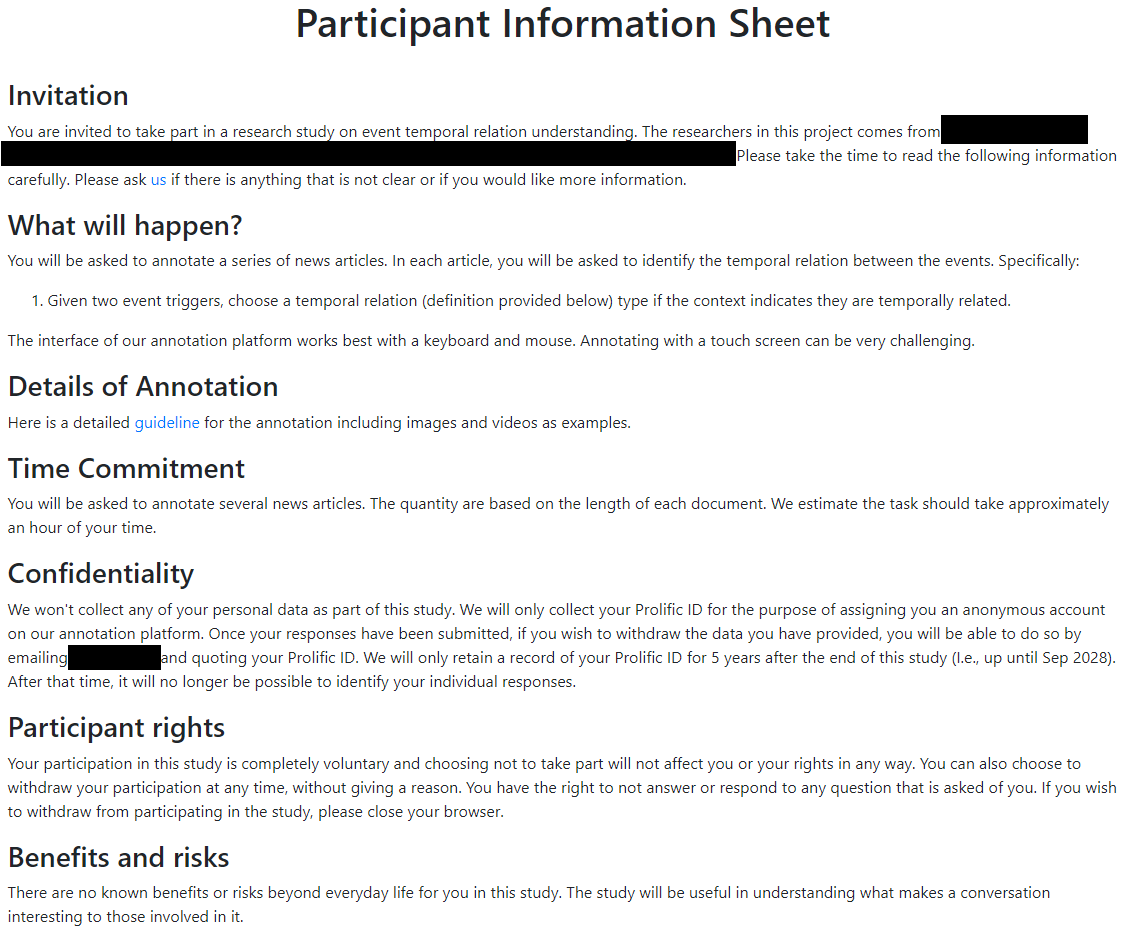}
   \caption{Disclaimers.}
   \label{fig:disclaimers}
\end{figure*}

\begin{figure*}[h] % 'h' for here
   \centering
   \includegraphics[width=0.95\textwidth]{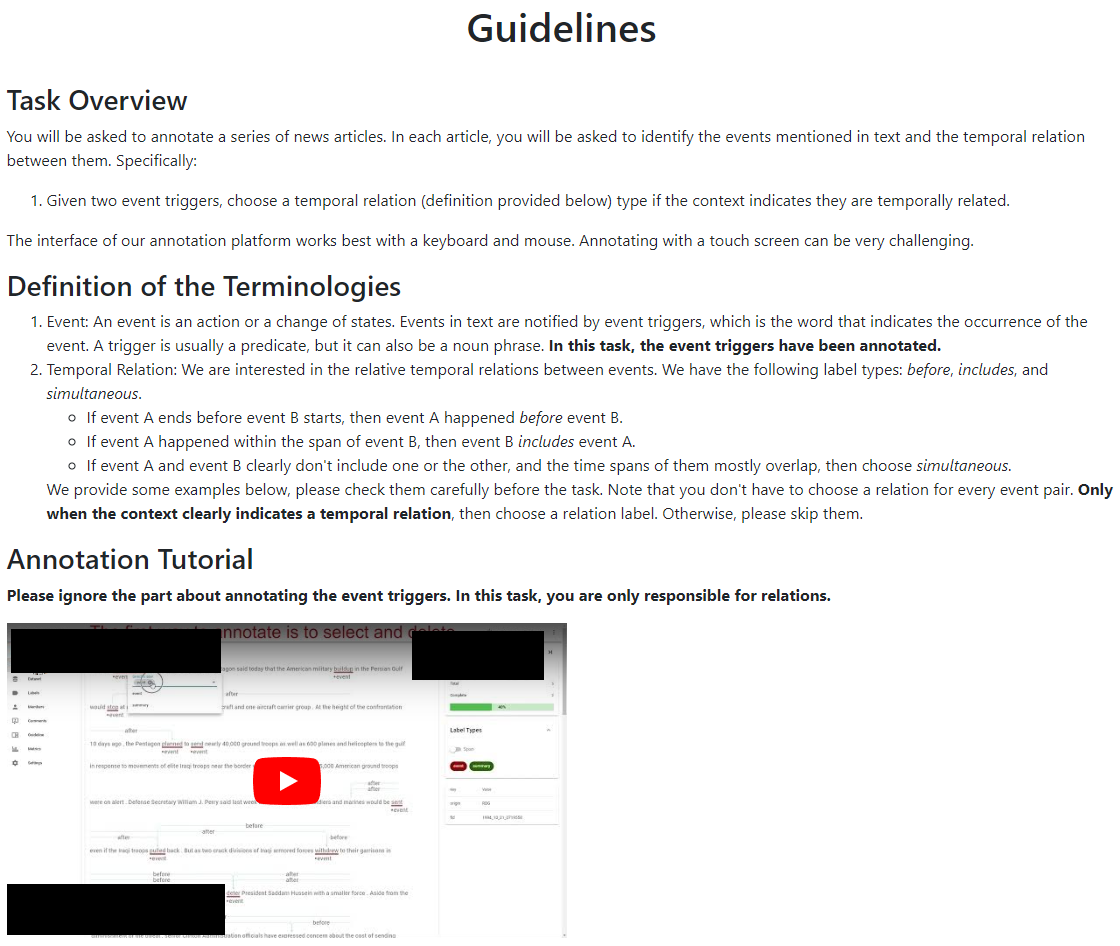}
   \caption{Guides.}
   \label{fig:guides}
\end{figure*}

\begin{figure*}
    
\begin{lstlisting}
FUNCTION_LIST = [
    {
        "name": "save_events",
        "description": "Store the extracted events in a list",
        "parameters": {
            "type": "object",
            "properties": {
                "event_list": {
                    "type": "string",
                    "description": "This is a list of event strings",
                }
            },
            "required": ["event_list"],
        },
    },
    {
        "name": "save_graph",
        "description": "Store the constructed graph in DOT language",
        "parameters": {
            "type": "object",
            "properties": {
                "graph": {
                    "type": "string",
                    "description": "The constructed graph in DOT language. \
                    This graph is a strict graph, in which every edge containing \
                    two event nodes, and a temporal relation label from \
                    [\"before\", \"includes\", \"simultaneous\"]. For example, \
                    \"strict graph  {\n\"The Organization asserted responsibility \
                   \" -- \"a United States Navy diver killed\"  [rel=before];\n}\"",
                }
            },
            "required": ["graph"],
        },
    }
]
\end{lstlisting}
\caption{Function list for OpenAI API call}
\label{fig:function_call}

\end{figure*}

\begin{table*}[t]
  \begin{center}
  \resizebox{\textwidth}{!}{%
  \begin{tabularx}{\textwidth}{X|X}
  \toprule
  Input & Output \\ 
  \midrule 
  First hop:
  
  Instructions: Analyse the given document. Extract the main events. Store the events as a JSON list by calling the save\_events function. 
  
  Example input document: Governor Cuomo leaving his office in Albany yesterday to speak to reporters after he met with representatives of group pushing for higher ethical standards for public officials in the state. 
  
  Example output events: Governor Cuomo leaving his office, Governor Cuomo speak to reporters, Governor Cuomo met with representatives."
  
  Document: President Clinton says he is committed to a possible strike against Iraq, despite objections from other countries. Steve Hurst has that. The US military buildup in the Persian gulf continues apace, more planes headed from the united states and senior officials say Iraq's president Saddam Hussein can expect punishing air strikes to continue well into the future if he doesn't stop building biological and chemical weapons. Saddam will know by our actions and our warning that we will be prepared to act again if we have evidence he is trying to rebuild his weapons of mass destruction capabilities. In a bit of television diplomacy, Iraq's deputy foreign minister responded from Baghdad in less than one hour, saying Washington would break international law by attacking without UN approval. The united states is not authorized to use force before going to the council. President Clinton, meantime, glossed over stern warnings from Moscow on Thursday that US air strikes against Iraq could do serious harm to relations with the Kremlin. We're trying to find a diplomatic solution. And I hope that, whatever happens today, that our relationships with Russia will continue to be productive and constructive and strong, because that's very important to the future of our peoples. One contrary view of the issue presented itself to the president as he arrived in Philadelphia later in the day. Nevertheless, the president said Washington would use force if diplomacy fails to force Saddam Hussein to back down.
  & strict graph \{
  
    "President Clinton's commitment to a possible strike against Iraq" -> "The US military buildup in the Persian Gulf" [label="before"];
    
    "The US military buildup in the Persian Gulf" -> "Senior officials' statement about punishing air strikes against Iraq" [label="before"];
    
    "Senior officials' statement about punishing air strikes against Iraq" -> "Iraq's deputy foreign minister's response from Baghdad" [label="before"];
    
    "President Clinton's attempt to find a diplomatic solution" -> "Contrary views expressed in Philadelphia" [label="before"];

    "Contrary views expressed in Philadelphia" -> "President Clinton's statement about using force if diplomacy fails" [label="before"];
    
    "The Russian foreign minister's suggestion to wait for UN Secretary General Kofi Annan's visit" -> "Secretary of State Madeleine Albright, Berger, and Defense Secretary William Cohen's plans to travel and explain military force necessity" [label="before"];

    \}

\\
\end{tabularx}}
\end{center}
  
\end{table*}
%%%%% example log of API call inputs and outputs

\begin{table*}[t]
  \begin{center}
  \resizebox{\textwidth}{!}{%
  \begin{tabularx}{\textwidth}{X|X}

 The Russian foreign minister, meanwhile, sought to soften the harsh words of his military counterpart, saying on Friday that Russia now feels the US must hold off at least until UN secretary general Kofi Annan visits Baghdad in a last-ditch effort at diplomacy. Annan has no trip planned so far. Meanwhile, secretary of state Madeleine Albright, Berger and defense secretary William Cohen announced plans to travel to an unnamed city in the US heartland next week, to explain to the American people just why military force will be necessary if diplomacy fails. Steve Hurst, CNN, the state department.

  Second hop:

    For every possible pair of events, please infer the relative temporal order (e.g., before, includes, simultaneous) between them based on the context. Make a graph where the nodes are the events, and the edges represent the temporal relation between the events. Lastly, please output the graph using DOT graph description language.

Conversation link: 

\url{https://chat.openai.com/share/a7c8659d-4fbd-4cfa-b60c-b4066ac77ee9}
  
  & 

\\
  \bottomrule
  \end{tabularx}}
  \end{center}
\end{table*}

\begin{table*}[t]
  \begin{center}
  \resizebox{\textwidth}{!}{%
  \begin{tabularx}{\textwidth}{X|X}
\toprule
  First hop:

Instructions: Analyse the given document. Extract the main events. Store the events as a JSON list by calling the save\textunderscore events function.
  
  Document: For The World, this is Pamela McCall in the BBC news room in London. More heavy gunfire in the Serbian province of Kosovo. It's the second day of an offensive by the security forces against villages populated by ethnic Albanians. BBC correspondent Karyn Coleman reports from Kosovo. There were more accounts of people fleeing from the villages near where the offensive was taking place. They were too afraid to stay, fearing the forces may also move their operations further afield. By mid afternoon, official Serb sources were saying the operation was over, but that has not yet been confirmed from Belgrade, the capital of Serbia, which is where the whole attack is thought to have been planned. The ethnic Albanians here in Kosovo are considerably worried about the future and fear they are witnessing the beginnings of a much wider conflict. Karyn Coleman, BBC news, Pristina.

  Second hop:

    For every possible pair of events, please infer the relative temporal order (e.g., before, includes, simultaneous) between them based on the context. Make a graph where the nodes are the events, and the edges represent the temporal relation between the events. Lastly, please output the graph using DOT graph description language.

    Conversation link: 

\url{https://chat.openai.com/share/6bdeb237-247d-486f-91ba-146c3b4fd3b2}

  &
strict graph \{

"Heavy gunfire in the Serbian province of Kosovo." -- "The second day of an offensive by the security forces against villages populated by ethnic Albanians" [rel=before];

    "The second day of an offensive by the security forces against villages populated by ethnic Albanians" -- "People fleeing from villages near the offensive" [rel=before];
    
    "The second day of an offensive by the security forces against villages populated by ethnic Albanians" -- "Official Serb sources stating the operation was over" [rel=before];
    
    "Official Serb sources stating the operation was over" -- "Uncertainty regarding the operation's status in Belgrade" [rel=simultaneous];
    
    "Uncertainty regarding the operation's status in Belgrade" -- "Ethnic Albanians in Kosovo expressing worry about the future and fear of a wider conflict" [rel=before];

\}
\\
  \bottomrule
  \end{tabularx}}
  \end{center}
  \caption{Example of ChatGPT promoting on the TBD data.}
  \label{prompt_table}
\end{table*}

\end{document}